\documentclass[10pt, final, journal, letterpaper, oneside, twocolumn]{IEEEtran}
\IEEEoverridecommandlockouts

\usepackage{graphicx} 

\usepackage{cite}
\usepackage{amsmath,amssymb,amsfonts}
\usepackage{algorithmic}
\usepackage{graphicx}
\usepackage{textcomp}
\usepackage{xcolor}
\usepackage{enumitem}
\usepackage{booktabs}
\usepackage{caption}
\usepackage{multirow}
\usepackage{makecell}
\usepackage{subcaption}
\usepackage[hidelinks]{hyperref}
\usepackage{hyperref}
\title{
EcoFollower: An Environment-Friendly Car Following Model Considering Fuel Consumption
}

\author{
    Hui Zhong,
    Xianda Chen,
    PakHin Tiu,
    Hongliang Lu,
    Meixin Zhu

\thanks{
    (\textit{Corresponding author: Meixin Zhu}) Hui Zhong, Xianda Chen and Meixin Zhu are with the Intelligent Transportation Thrust, The Hong Kong University of Science and Technology (Guangzhou), Guangzhou, 511400, China; Meixin Zhu is also with Guangdong Provincial Key Lab of Integrated Communication, Sensing and Computation for Ubiquitous Internet of Things.
    This study is supported by the National Natural Science Foundation of China under Grant 52302379 and 62373315, Guangzhou Basic and Applied Basic Research Projects under Grants 2023A03J0106, 2023A03J0683 and 2024A04J4290, Guangdong Province General Universities Youth Innovative Talents Project under Grant 2023KQNCX100, and Guangzhou Municipal Science and Technology Project 2023A03J0011.(email: hzhong638@connect.hkust-gz.edu.cn,
    xchen595@connect.hkust-gz.edu.cn, phtiu454@connect.hkust-gz.edu.cn, meixin@ust.hk)
}
}

\begin{document}

\maketitle
\thispagestyle{empty}
\pagestyle{empty}

\begin{abstract}
To alleviate energy shortages and environmental impacts caused by transportation, this study introduces EcoFollower, a novel eco-car-following model developed using reinforcement learning (RL) to optimize fuel consumption in car-following scenarios. Employing the NGSIM datasets, the performance of EcoFollower was assessed in comparison with the well-established Intelligent Driver Model (IDM). The findings demonstrate that EcoFollower excels in simulating realistic driving behaviors, maintaining smooth vehicle operations, and closely matching the ground truth metrics of time-to-collision (TTC), headway, and comfort. Notably, the model achieved a significant reduction in fuel consumption, lowering it by 10.42\% compared to actual driving scenarios. These results underscore the capability of RL-based models like EcoFollower to enhance autonomous vehicle algorithms, promoting safer and more energy-efficient driving strategies.

\end{abstract}

\section{INTRODUCTION}
The escalating concerns over global energy shortages and environmental issues have driven increased focus on emission control in autonomous vehicles \cite{dewees1989woodfuel}. The transportation sector is a major contributor, accounting for a staggering 59\% of total oil consumption and 22\% of carbon dioxide emissions in 2011 \cite{/content/publication/co2_fuel-2013-en}.  Furthermore, transportation was responsible for nearly 30\% of all greenhouse gas (GHG) emissions in 2015, highlighting its role in global warming and its detrimental effects \cite{salvi2015experimental}.


Car-following behaviour is a fundamental aspect of autonomous driving systems, particularly within Adaptive Cruise Control (ACC) frameworks  \cite{chen2024aggfollower, chen2024genfollower}. It allows autonomous vehicles to maintain a safe distance from preceding vehicles by adjusting their speed \cite{gipps1981behavioural}. Traditionally, car-following models have been built using two main approaches: rule-based and supervised learning \cite{saifuzzaman2014incorporating}. Rule-based models include classics like the Gazis-Herman-Rothery (GHR) model and the Intelligent Driver Model (IDM) \cite{gazis1961nonlinear}. Supervised learning models, on the other hand, rely on data from human drivers to learn the relationship between car-following states and vehicle acceleration \cite{chen2024editfollower, zhu2022transfollower, chen2024metafollower}. However, both approaches aim to replicate human driving behaviour. Studies have shown that novice drivers tend to be less smooth or more aggressive with the accelerator pedal compared to experienced drivers, leading to higher fuel consumption \cite{huang2021impact}. This suggests that simply mimicking human behaviour might not be the most optimal solution for autonomous driving. Additionally, research on fuel economy testing of autonomous vehicles has indicated that algorithms not designed with efficiency in mind can lead to a 3\% decrease in fuel economy \cite{mersky2016fuel}. These findings emphasize the importance of considering environmental impact and implementing sustainable traffic management strategies.

To tackle these challenges, this paper introduces a groundbreaking human-like eco-car-following model based on reinforcement learning (RL), named EcoFollower. This model distinguishes itself by incorporating passenger comfort alongside safety, efficiency, and fuel consumption optimization. We evaluate the effectiveness of the proposed model using real-world data, focusing on its ability to improve fuel efficiency while maintaining passenger comfort demonstrably. This analysis of real-world data allows us to demonstrate the Ecofollower's capacity to enhance energy efficiency in autonomous driving systems significantly.





\section{RELATED WORK}

The growing interest in autonomous driving stems partly from its potential to reduce energy consumption. Studies have explored this potential using two main approaches: macro-level simulations and individual vehicle behaviour optimization.

Chen $et$ $al.$ \cite{chen2019quantifying} employed network activity data to simulate national fuel consumption impacts in the U.S., offering insights into the broader implications of autonomous driving. Similarly, Yao $et$ $al.$ \cite{yao2021fuel} compared fuel consumption and emissions between connected automated vehicles and human-driven vehicles, highlighting the potential environmental benefits. However, these valuable insights are limited to a macro level, and there is a need to investigate individual vehicle behaviors within the context of autonomous driving.

Wu et al. \cite{wu2011fuel} proposed an optimization system for improving fuel economy by fine-tuning rule-based car-following models. However, research on supervised learning approaches, which are becoming increasingly prevalent, remains relatively scarce. Zhou et al. \cite{zhou2019development} introduced a driving strategy for connected and automated vehicles at signalized intersections using a reinforcement learning (RL) approach, demonstrating its effectiveness in macro-level traffic management. While these studies address aspects of safety, efficiency, and comfort, most car-following models tend to overlook fuel economy as a primary objective. Zhu et al. \cite{zhu2020safe} showed that RL-based car-following models outperform Model Predictive Control (MPC) based ACC algorithms in terms of balancing objectives like safety, efficiency, and comfort. This highlights the potential of RL for developing car-following models that achieve high performance in these combined areas.

These works reveal a gap in research on car-following models that simultaneously consider fuel efficiency and passenger comfort.  Furthermore, the success of RL-based approaches in achieving a balance between various objectives suggests its potential for developing a new model that addresses this gap. Our proposed EcoFollower model is designed to bridge this gap by incorporating passenger comfort alongside safety, efficiency, and fuel consumption optimization through an RL-based approach.

\section{DATA PREPARATION}

Building upon our previous work on car-following datasets in \cite{chen2023follownet}, we selected the Next Generation Simulation (NGSIM) dataset \cite{NGSIM-DATASET} for this study due to its well-established reputation and alignment with the criteria identified in our earlier research. The data encompasses a 500-meter section with six freeway lanes, including a dedicated high-occupancy vehicle (HOV) lane.  It provides a total of 45 minutes of data divided into three 15-minute intervals (4:00 PM - 4:15 PM, 5:00 PM - 5:15 PM, and 5:15 PM - 5:30 PM). These specific periods capture the transition from uncongested to congested traffic and peak rush hour conditions. The dataset records each vehicle's location at a high sampling rate of 10 Hz, ensuring precise data. We employed the reconstructed NGSIM I-80 data \cite{montanino2015trajectory} to guarantee data quality. A total of 1,341 car-following events that are over 15 seconds were extracted from the dataset for this study.


\begin{figure}
  \centering
  \includegraphics[width=0.45\textwidth]{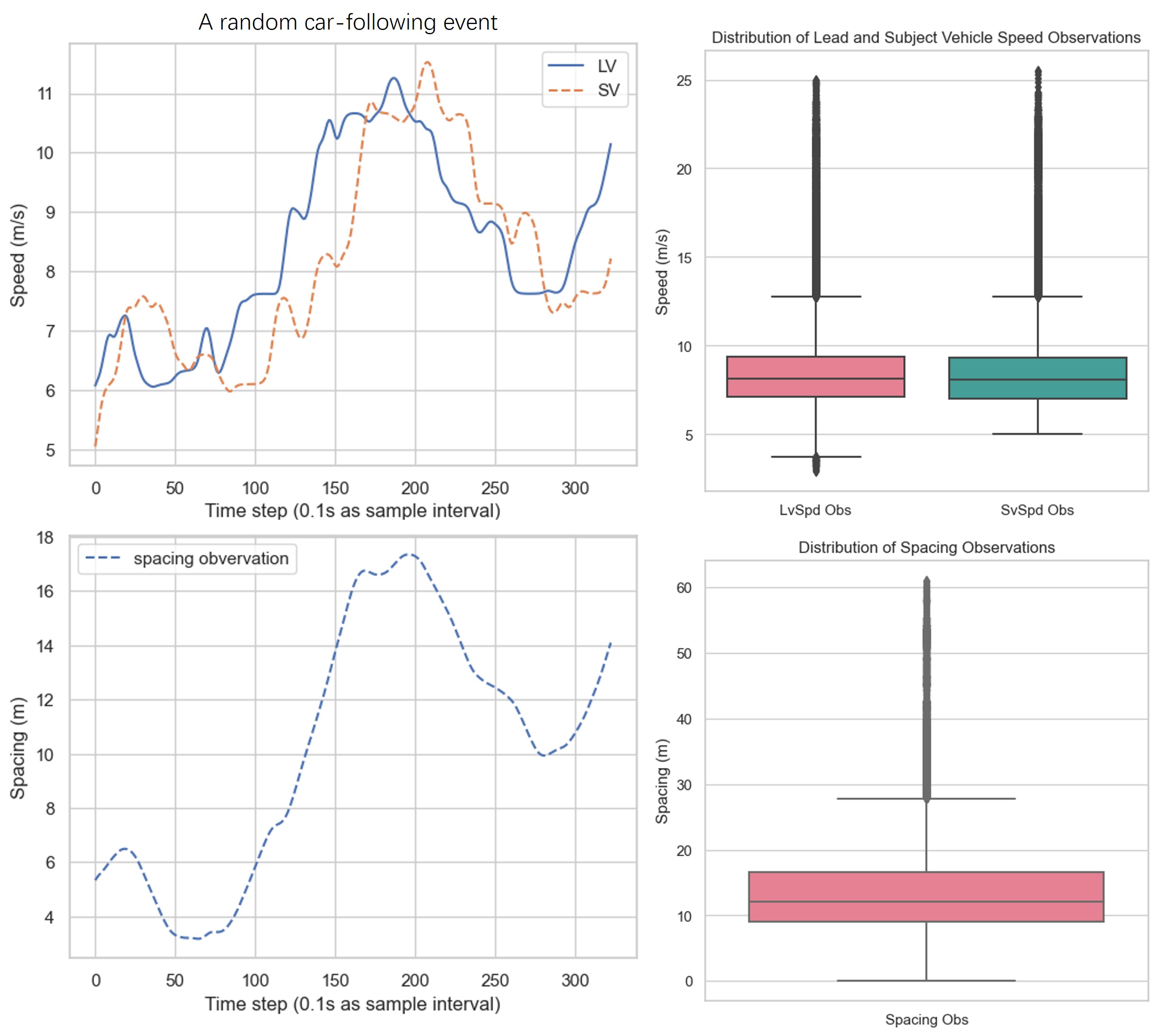} 
  \caption{A random car-following event and descriptive analysis} 
  \label{fig.1} 
\end{figure}

Fig.\ref{fig.1} illustrates the statistic analysis of the NGSIM dataset, revealing insights into following vehicle behavior. The data suggests generally smooth traffic flow, with average speeds for leading and following vehicles at 8.14 m/s and 8.07 m/s, respectively. Additionally, the average distance headway is 12.12 m, with a maximum of 60 m. This suggests that following vehicles typically maintain close proximity without experiencing excessive braking. However, Figure \ref{fig.1} also highlights the presence of random following events characterized by significant variations in speed and headway. While these events are brief, lasting only around 30 seconds, they involve rapid acceleration and deceleration. This suggests potential discomfort for passengers in following vehicles due to unexpected changes in speed.

\section{METHODOLOGY}

\subsection{Intelligent Driver Model (IDM)}

IDM serves as our baseline model for car-following behavior.  IDM is a well-established model that mimics human driving by considering both the desired speed of the following vehicle and its desired following distance. Mathematically, IDM is expressed as follows:
\begin{align}
    a_{n}(t) = a_{\text{max}}[1-(\frac{v_{n}(t)}{\tilde{v_{n}(t)}})^\beta -(\frac{\tilde{s_{n}(t)}}{s_{n}(t)})^2]
    \label{eq:idm}
\end{align}
where $a_{n}(t)$ is the output acceleration for the following vehicle $n$, $a_{max}^{(n)}$ is the maximum acceleration, $v_{n}(t)$ and $\tilde{v_{n}(t)}$ are the actual and desired velocity, respectively, and $s_{n}(t)$ and $\tilde{s_{n}(t)}$ are the actual and desired spacing, respectively. Specifically, the desired spacing can be calculated by:
\begin{align}
    \tilde{s_{n}(t)} = s_{\text{jam}}^{(n)} + \max \left(0, v_{n}(t) T_{n} + \frac{v_{n}(t) {\Delta v}_{n}(t)}{2 \sqrt{a_{\text{max}} a_{\text{comf}}}}\right)
    \label{eq:desired-spacing}
\end{align}
where $s_{\text{jam}}^{(n)}$ is the minimum standstill spacing, ${T}_{n}$ is the desired time headway, ${\Delta v}_{n}(t)$ is the relative speed between the leading vehicle and the following vehicle $n$, and $a_{\text{comf}}$ is the comfortable acceleration. The specific model parameters used in this study were calibrated based on the work of Zhu $et$ $al.$ \cite{zhu2020safe}.


\subsection{DDPG Algorithm}

This section explores the Deep Deterministic Policy Gradient (DDPG) algorithm, employed to train an intelligent agent that simulates the following vehicle's behavior. The goal is to develop an optimal policy that achieves a balance between fuel consumption, safety, travel effectiveness, and passenger comfort in car-following scenarios.

While our primary focus is on evaluating the potential of DDPG for fuel-efficient car following, it's well established that energy consumption is directly related to vehicle speed and acceleration. However, simply minimizing fuel consumption might lead the agent to prioritize staying stationary, compromising travel efficiency. Therefore, we design a reward function that considers a combination of factors: fuel consumption, time to collision (TTC), time headway, and jerk. This approach balances fuel efficiency with safety and travel progress. The specific parameters used in the reward function will be detailed in the following section.



\subsubsection{Fuel Consumption}

The VT-Micro model, established by Ahn et al. in 2002 \cite{ahn2002estimating}, serves as the foundation for estimating fuel consumption in the following vehicle. This well-regarded model is known for its high accuracy and has been validated with real-world data. Its strength lies in considering both speed and acceleration through an exponential Measure of Effectiveness (MOE) \cite{zhou2023experimental}. The MOE is expressed mathematically as:
\begin{align}
    MOE(a_{n}(t), v_{n}(t)) = \exp{P(a_{n}(t), v_{n}(t))}
    \label{eq:MOE}
\end{align}
where is $P(a_{n}(t), v_{n}(t))$ a polynomial function representing the influence of speed and acceleration on fuel consumption, represented as:
\begin{align}
    P(a_{n}(t), v_{n}(t)) = \sum_{i=0}^{3} \sum_{j=0}^{3} K_{ij} v_{n}^i a_{n}(t)^j
    \label{eq:fuel-influence}
\end{align}
where $K_{ij}$ are the regression coefficients that have been calibrated using field data, and they quantify the influence of speed and acceleration on fuel consumption, $CO_2$ emissions, and $NO_x$ emissions.

\subsubsection{Comfort}

We leverage jerk, the rate of change of acceleration, as the primary metric for quantifying passenger comfort. Jerk represents the jolting sensation passengers experience during abrupt changes in speed. Minimizing jerk translates to smoother vehicle motion and reduces discomfort. Jerk can be calculated using the following equation:
\begin{align}
    J_{n}(t) = \frac{d}{dt} a_{n}(t)
\end{align}
where $J_{n}(t)$ and $a_{n}(t)$ are the jerk and acceleration of the following vehicle $n$ at time $t$.

\subsubsection{Safety and Efficiency}
According to previous studies\cite{zhu2020safe, pu2020full, han2023ensemblefollower}, safety and efficiency can be denoted by the TTC and time headway. Safety is the most essential and important element in traffic conditions and 
TTC means the time left before a collision takes place during a car-following event. Thus, it can be represented using eq.\ref{eq.2}.
\begin{equation}
    TTC(t) = -\frac{S_{{n-1},n}(t)}{\Delta V_{{n-1},n}(t)}
    \label{eq.2}
\end{equation}
\\where $t$ denotes time step; $n-1$ and $n$ stands for the preceding and following vehicle, respectively. Therefore, $S_{{n-1},n}(t)$ represents the spaces between two vehicles at moment $t$, m; $\Delta V_{{n-1},n}(t)$ means the relative speed at moment $t$, which equals to the speed of leading vehicles minus the speed of following vehicles. This definition reflects that the trend of TTC is converse with crash risk \cite{vogel2003comparison}. Usually, the thresholds of TTC range from 1.5 s to 5s. In this study, its thresholds were referred to the results of Zhu et al. \cite{zhu2020safe}, which can be constructed as follows.
\begin{equation}
    F_{TTC}=\left\{\begin{matrix}
log(\frac{TTC}{4}), & 0\leq TTC\leq 4\\ 
 0, & otherwise 
\end{matrix}\right.
\end{equation}
\\Time headway refers to the duration between the arrival of the leading vehicle and the following vehicle at a specific point. In general, a shorter time headway signifies a more efficient transportation condition characterized by larger vehicle capacities \cite{zhang2007examining}. Generally, the detailed calculation can be expressed as follows:
\begin{equation}
     h_i = \frac{x_{i-1}(t) - x_i(t)}{v_i(t)}
\end{equation}
Here, $x_{i-1}(t)$ and $x_i(t)$ denote the positions of the leading vehicle$i-1$ and the following vehicle $i$ at time $t$, respectively, while $v_i(t)$ represents the speed of the following vehicle $i$ at time $t$
According to the distribution of time headway, the probability density function can be constructed as a lognormal function. Therefore, we can establish the probability density function of the lognormal distribution, where $x$ is the variable of time headway. $\mu$ and $\sigma$ is the mean and log standard deviation of the time headway ($x$). Based on the empirical 1341 events of NGSIM, the calculated $\mu$ and $\sigma$ are 0.4226 and 0.5436, respectively. 
\\Fig. 2 presents a descriptive analysis of four indicators derived from the dataset used in the test set. The entire dataset was divided into training and test sets in a 7:3 ratio, with the latter employed to assess the performance of various models. The distribution of the time-to-collision (TTC) in the raw data is relatively balanced, ranging from -20.76 to 21.51 seconds. This distribution indicates that close to 50$\%$ of the following events in the data carry a risk of collision, with some having already resulted in an impact. The comfort of driving is typically evaluated by the rate of change of acceleration, known as a jerk. A comfortable driving experience is generally characterized by jerk values remaining within 1 m/s³, signifying that the changes in acceleration are not abrupt and the vehicle motion is comparatively smooth, offering passengers a ride without significant jolts or discomfort. The jerk values in the dataset mostly range between -0.36 and 0.39 $m/s^3$, indicating gentle changes in acceleration and a comfortable driving style. 

\begin{figure}[h]
  \centering
  \includegraphics[width=0.5\textwidth]{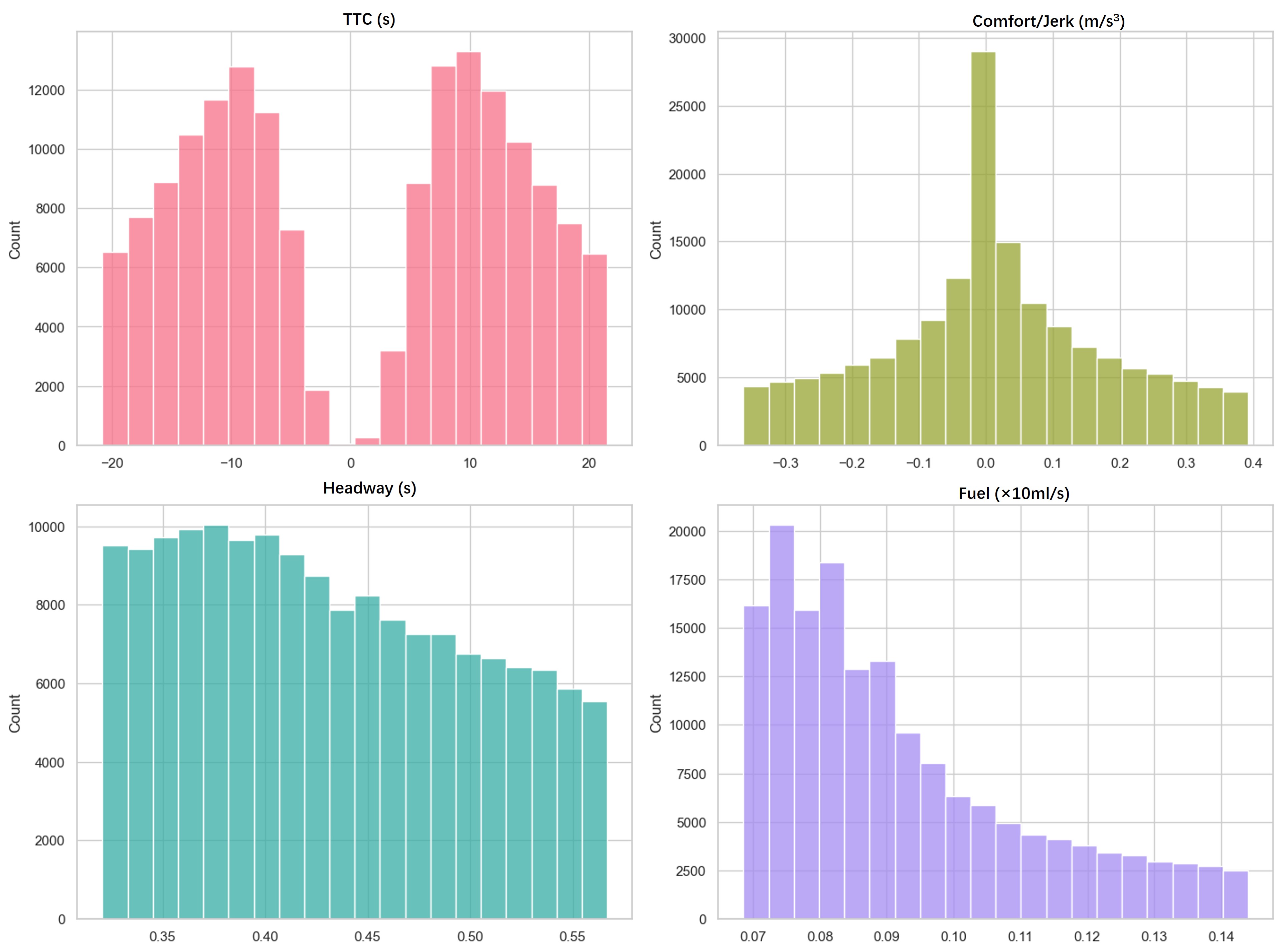} 
  \caption{descriptive analysis of four indicators} 
  \label{fig.2} 
\end{figure}

\begin{equation}
\begin{matrix}
f_{lognorm}(x|\mu , \sigma )=\frac{1}{x\sigma \sqrt{2\pi }}e^{-\frac{(lnx-\mu )^2}{2\sigma ^2}}, & x>0
\end{matrix}
\label{eq.4}
\end{equation}

\subsubsection{States and Actions}
In the car-following scenario, the action of the agent was defined as the longitudinal acceleration of followers, and the state contains speed $V_{FV}(t)$, relative gap $S(t)$ and relative speed $\Delta V{LV, FV}$ of the following vehicles. Thus, at each time step $t$, the simulated follower's trajectory satisfied the Newtonian motion equations given lead vehicle data, which can be divided into these three steps:
\begin{itemize}
    \item A. Calculate action (acceleration)
    \item B. Update state (speed, relative speed and spacing):\\


\begin{flalign}\label{eq:update}
\begin{split}
&\Delta V(t+1)=V_{LV}(t+1)-V_{FV}(t+1)\\
&S(t+1)=S(t)+\frac{\Delta V(t)+\Delta V(t+1)}{2}*\Delta T
\end{split}
\end{flalign}

    \item Return to Step A until the event ends
\end{itemize}

\subsubsection{Reward Function}
According to the above-mentioned features for reward function, it can be defined as the following equation, which considers safety and fuel consumption and is subject to traffic efficiency.
\begin{equation}
    r(s, a) = \omega_1 F_{TTC}+\omega_2 F_{headway}+\omega_3 F_{fuel}+ \omega_4 F_{jerk}
\end{equation}
\\where $r(s, a)$ denotes the reward function and $\omega_1, \omega_2, \omega_3$ stand for weights of safety, efficiency, and fuel consumption, respectively, during car following. In this paper, they are set to 1 given for equal importance\cite{zhu2020safe}.

\section{EXPERIMENT}
\subsection{EcoFollower model}
The NGSIM datasets are divided into training and testing subsets, utilizing 70$\%$ of the data for training and 30$\%$ for testing purposes. During the training phase, the RL agent iteratively simulates car-following events using the training dataset, which undergoes random shuffling. After the completion of each event, a new car-following event is randomly drawn from the training dataset, and the agent’s state is reset based on the empirical data from the selected event. This iterative training cycle is conducted across 3000 episodes, with each episode corresponding to a unique car-following scenario.

\begin{figure*}
  \centering
  \includegraphics[width=1\textwidth]{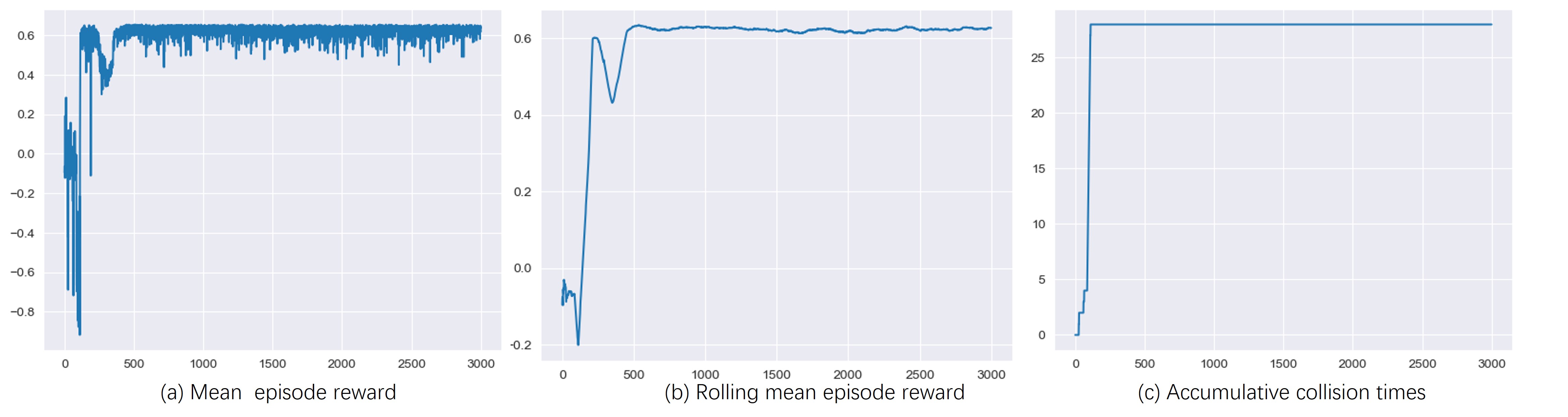} 
  \caption{The training log of reward and collision } 
  \label{fig.3} 
\end{figure*}

Fig.\ref{fig.3} depicts the training outcomes after 3000 episodes, clearly demonstrating that the DDPG model achieves convergence by approximately the 500th episode when employing a collision avoidance strategy. The rolling mean episode reward, calculated as the average reward obtained across all time steps within an event at a sampling interval of 0.1 seconds for the NGSIM data, stabilizes around 0.6, accompanied by 28 accumulative collisions. These findings suggest that the EcoFollower car-following model is sufficiently adaptable to develop an optimal strategy that effectively balances safety, efficiency, comfort, and fuel consumption.

Fig.4 presents the test results for safety, comfort, efficiency, and fuel consumption using the EcoFollower car-following model. When compared to Figure 2, the distributions of the four calculated indicators in the test results mirror the trends observed in the original training datasets. For instance, the analysis of the original test datasets indicated that the TTC values ranged from -20 seconds to 20 seconds. The predictive outcomes of TTC from the trained EcoFollower model exhibit a symmetrical range. Likewise, the distribution of jerk maintains a consistent pattern between the observed and simulated values, spanning the range of [-0.4, 0.4]. These observations underscore the substantial simulation capabilities of the DDPG model in replicating human driver behaviours, as noted in the work of Zhu et al.\cite{zhu2018human}. Moreover, a comparison of fuel consumption between real and simulated car-following events reveals an approximate 10.42$\%$ energy savings using the EcoFollower model, with a mean fuel consumption rate of 0.86ml/s across 403 car-following events. This demonstrates the model's efficacy not only in mimicking driving behaviours but also in enhancing fuel efficiency in practical scenarios. 
\begin{figure}[h]
  \centering
  \includegraphics[width=0.49\textwidth]{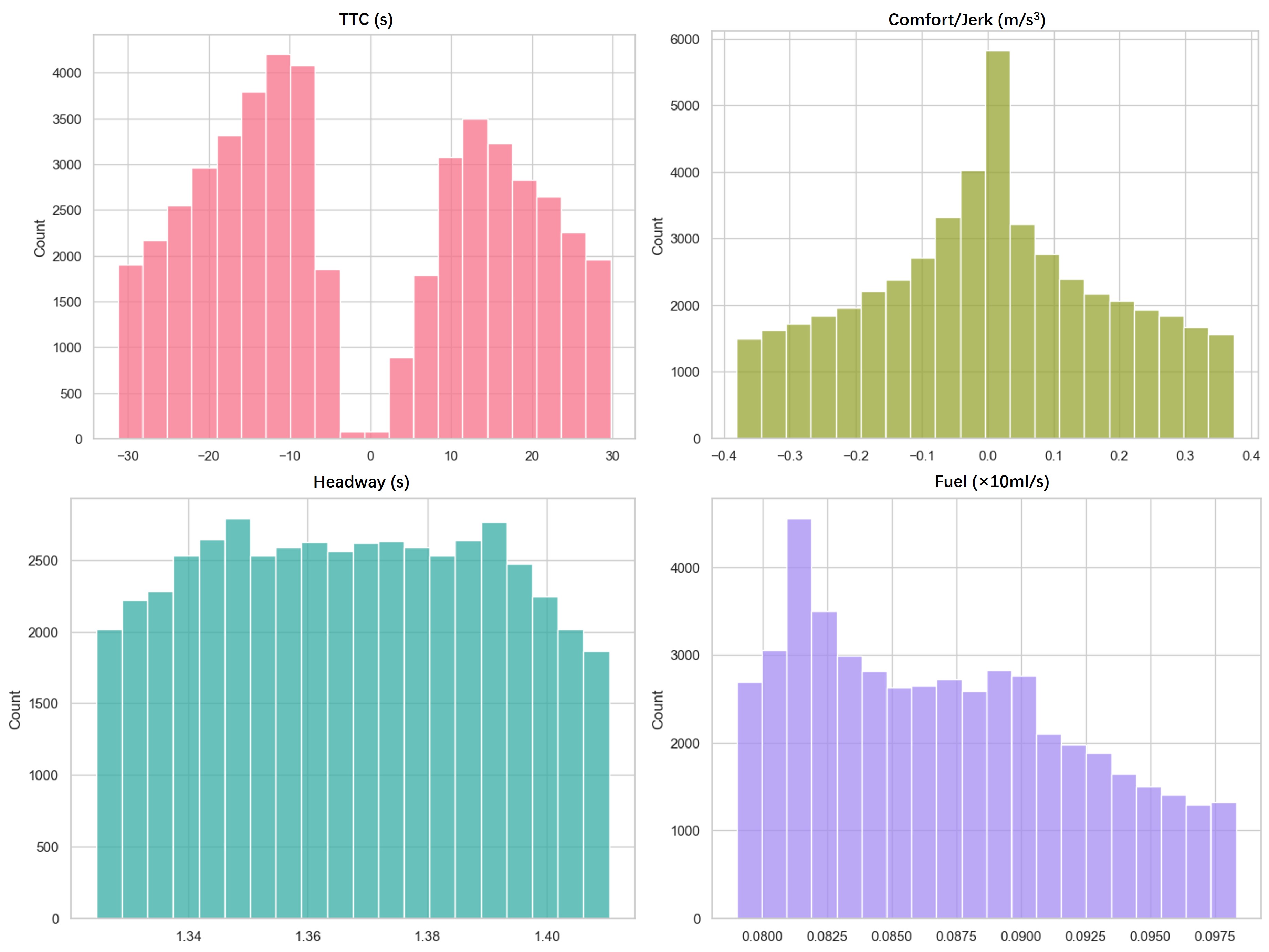} 
  \caption{The test results of safety, comfort, efficiency and fuel consumption in EcoFollower model} 
  \label{fig.4} 
\end{figure}

\subsection{Comparison between EcoFollower and IDM model}

To evaluate the efficacy of RL in reducing fuel consumption within autonomous driving contexts, this study conducted comparative experiments using the trained car-following model and the IDM on the same test datasets, assessing performance in terms of safety, comfort, efficiency, and fuel consumption. Notable differences were observed in the simulated driving behaviours between the IDM and the RL-based models. The EcoFollower model demonstrated a high capacity for emulation, as evidenced by the similarity in the distribution of the four key indicators, whereas the driving behaviours modelled by the IDM exhibited considerable variability from real conditions.

Furthermore, the safety of simulated car-following behaviours by the IDM model showed significant enhancements, characterized by larger TTC values and increased headway. Conversely, comfort levels decreased slightly in comparison to real-world conditions, with a broader jerk range of [-0.6, 0.8]. Additionally, the fuel consumption exhibited an upward trend due to frequent acceleration and deceleration. Prior studies have indicated that more frequent "stop-and-go" behaviours typically result in increased fuel consumption and emissions. This highlights the complex trade-offs between achieving safety and efficiency in autonomous driving technologies and underscores the importance of optimizing driving algorithms to balance these factors effectively.
\begin{figure}
  \centering
  \includegraphics[width=0.49\textwidth]{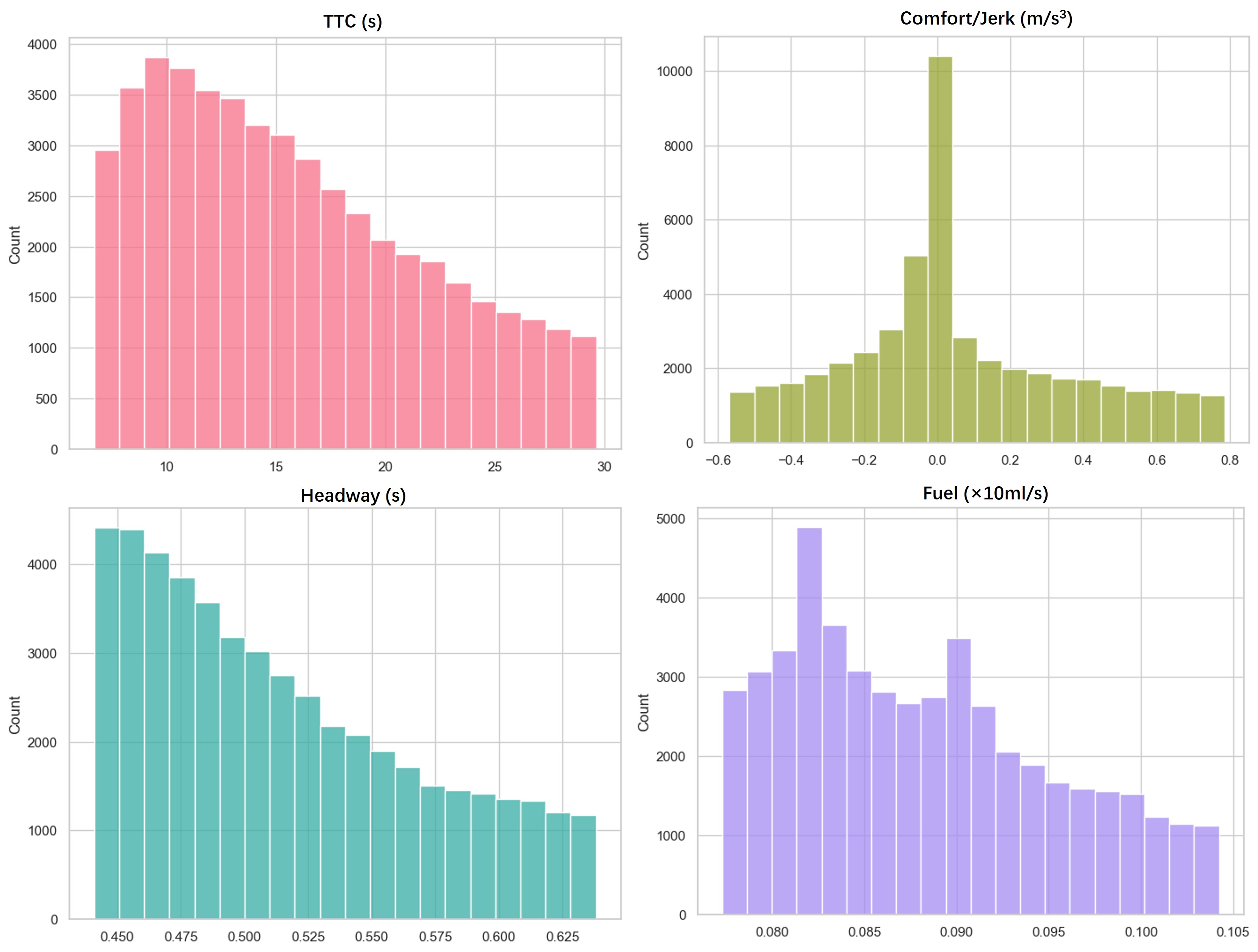} 
  \caption{The test results for safety, comfort, efficiency and fuel consumption in IDM model} 
  \label{fig.5} 
\end{figure}

Table 1 provides a comprehensive evaluation of safety, comfort, efficiency, and fuel consumption, comparing the simulated results with the ground truth. The findings indicate that the simulated car-following models generally adopt a conservative strategy to enhance safety, as evidenced by higher TTC values. However, among these models, the IDM demonstrates the poorest comfort levels but achieves the highest efficiency. This outcome is attributed to the frequent acceleration and deceleration behaviours exhibited by the simulated following vehicle, which, in turn, lead to relatively higher fuel consumption compared to the EcoFollower model. 


\begin{table*}[htp]
    \centering
    \begin{tabular}{c|c|c|c|c}
        \toprule
        Model & TTC ($s$) & Jerk ($m/s^{3}$) & Time Headway ($s$) & Fuel Consumption ($mL/s$) \\
        \midrule
        EcoFollower & 11.399 & 0.377 & 1.393 & 0.86 \\
        IDM & 14.788 & 1.187 & 0.503 & 0.87 \\
        Ground-truth & 4.625 & 0.676 & 1.476 & 0.96 \\
        \bottomrule
    \end{tabular}
    \caption{Four indicators for safety, comfort, efficiency and fuel consumption among different car-following model}
    \label{tab:my_label}
\end{table*}

To examine the variations among different models, four indicators were analyzed during a randomly selected car-following event, as presented in Figure 6. The actual data for TTC, headway, and fuel consumption exhibit significant fluctuations. To some extent, both the EcoFollower and IDM models mitigate these oscillatory fluctuations that occur during car-following. It is also noteworthy that, due to improper initial state settings, the IDM model experienced a sudden change in speed within the first few seconds, leading to intense acceleration. Additionally, both headway and fuel consumption rates underwent abrupt changes. These factors contribute to the IDM model's inferior performance compared to the EcoFollower model. In contrast, the car-following behaviours simulated by the EcoFollower model result in smoother driving actions. These results underscore the importance of parameter calibration in the IDM model to improve its predictive accuracy and overall performance. It is still a challenge to balance various performance metrics within autonomous driving simulations to optimize overall vehicle performance.

\begin{figure}
  \centering
  \includegraphics[width=0.5\textwidth]{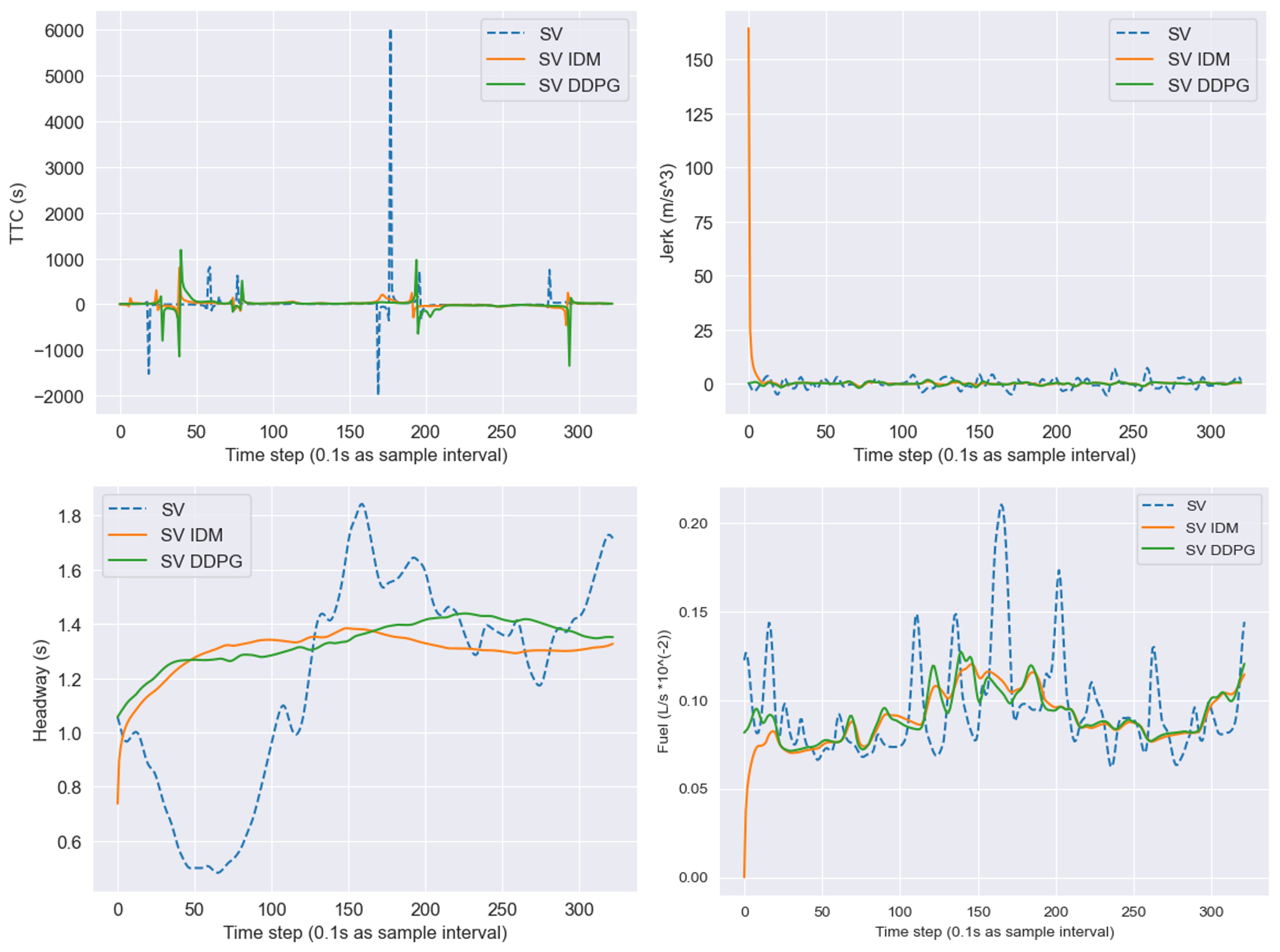} 
  \caption{Four indicators of safety, comfort, efficiency and fuel consumption in a random car-following event} 
  \label{fig.6} 
\end{figure}

\section{CONCLUSIONS}
This study developed and assessed the EcoFollower model, an innovative car-following model based on reinforcement learning designed to optimize fuel consumption. This model was contrasted with the traditional IDM  to evaluate comparative efficacy across key performance metrics, including safety, comfort, efficiency, and fuel consumption in both simulated and real-world settings. The results demonstrate that both the EcoFollower and IDM models are effective in mitigating the oscillatory fluctuations typical in car-following data. However, they exhibit distinct behavioral patterns that significantly affect their overall performance. The EcoFollower model not only excelled in reducing fuel consumption but also excelled in its ability to simulate realistic driving behaviors, maintain smooth vehicle operations, and closely mirror ground truth metrics such as TTC, headway, and jerk. Conversely, the IDM model, while demonstrating high efficiency with lower time headway, exhibited abrupt changes in speed and time headway in the first part of car-following events. These sudden alterations frequently led to increased fuel consumption and diminished comfort. These findings underscore the potential of advanced simulation models like EcoFollower to improve autonomous vehicle algorithms, leading to safer and more efficient driving strategies.




\bibliographystyle{ieeetr}

\bibliography{refs}

\end{document}